\documentclass[11pt]{article}
\usepackage{acl}
\usepackage{times}
\usepackage{latexsym}
\usepackage[T1]{fontenc}
\usepackage[utf8]{inputenc}
\usepackage{microtype}
\usepackage{inconsolata}
\usepackage{booktabs}
\usepackage{multirow}
\usepackage{graphicx}
\usepackage{amsmath}
\usepackage{placeins}
\usepackage{caption}
\usepackage{amssymb}

\title{Small LLMs for Biomedical Claim Verification: Cost-Effective Fine-Tuning, Structural Dataset Shortcuts, and Cross-Domain Generalization}

\author{Gaurav Kumar$^{1,2}$ \\
  $^1$Moveworks AI  \\
  $^2$University of California San Diego \\
  \texttt{gkumar@ucsd.edu}}

\begin{document}
\maketitle

\begin{abstract}
Large Language Models such as GPT-4o and GPT-5 achieve strong zero-shot
performance on biomedical claim verification, but cost and
opacity limit scalable use. We fine-tune three
small LLMs: Phi-3-mini (3.8B), Qwen2.5-3B, and
Mistral-7B, via QLoRA on SciFact and HealthVer, providing
the first study of QLoRA models
against GPT-4o and fine-tuned BioLinkBERT encoders. Mistral-7B QLoRA achieves higher F1 than both GPT-4o
and GPT-5 (up to 12\% gain) at 44.5$\times$
lower cost using just 1,008 training examples,
representing a compelling cost-quality trade-off. We conduct extensive in-domain and cross-domain evaluation:
models trained on SciFact tested on HealthVer and vice
versa, at matched sizes to isolate dataset
structure from data quantity.  We identify a previously unreported structural artifact in SciFact that inflates in-domain scores,
and show through bidirectional out-of-domain evaluation
that training on
structurally sound data enables robust cross-domain
transfer.
We release all code and adapter checkpoints.
\end{abstract}

\section{Introduction}

Automated biomedical claim verification determines whether a claim is \textsc{supported}, \textsc{refuted}, or undetermined by evidence---a form of Natural Language Inference
(NLI) applied to the biomedical domain. It is increasingly critical as health misinformation increases \cite{vladika2023scientific, guo2022survey, kotonya2020pubhealth}. Large language models (LLMs) such as GPT-4o offer strong zero-shot performance \cite{nori2023medprompt, kosprdic2024scientific}. However, API costs scale linearly with volume, they cannot be locally deployed in privacy-sensitive clinical environments, and silent provider updates undermine reproducibility. QLoRA \cite{dettmers2023qlora} enables 4-bit fine-tuning of billion-parameter models on a single GPU in under an hour, yet no systematic comparison of QLoRA-adapted small LLMs against both proprietary and encoder baselines exists for this task under conditions that jointly test in-domain performance and out-of-domain generalization.

We fine-tune Qwen2.5-3B \cite{qwen2025}, Phi-3-mini 3.8B \cite{abdin2024phi3}, and Mistral-7B-Instruct \cite{jiang2023mistral} on SciFact \cite{wadden-etal-2020-fact} and HealthVer \cite{sarrouti-etal-2021-evidence} separately, evaluating them bidirectionally alongside GPT-4o, GPT-5, and fine-tuned BioLinkBERT \cite{yasunaga2022linkbert}. Our contributions are:

\begin{enumerate}

\item QLoRA fine-tuning beats both GPT-4o and GPT-5
on biomedical claim verification at 44.5$\times$ lower
cost. Fine-tuning on just 1,008 examples enables
Mistral-7B to achieve 88.4\% macro-F1 on SciFact and
65.2\% on HealthVer, surpassing GPT-4o (85.6\%, 53.2\%)
and GPT-5 (77.9\%, 42.4\%) on both datasets, as well as earlier reported fine-tuned encoder results
\cite{kosprdic2024scientific}.

\item We identify a previously unreported
structural artifact in SciFact. All 243 NEI training
examples have empty evidence fields, making NEI trivially
detectable from evidence absence rather than genuine
reasoning. Fine-tuned models exploit this cue, achieving
100\% in-domain NEI F1, inflating macro-F1 in a way
unreported by prior work.

\item Bidirectional out-of-domain evaluation: Training
on SciFact and testing on HealthVer, and vice versa at
matched sizes demonstrates
robust cross-domain adaptability: fine-tuning
on just 1,008 HealthVer examples enables
Mistral-7B to achieve 74.3\% NEI F1 on SciFact OOD,
outperforming BioLinkBERT trained on 10$\times$ more
data (60.8\%). The reverse direction confirms the
shortcut mechanism: SciFact-trained models
collapse on HealthVer, with the
asymmetry ruling out domain shift and data quantity as
explanations.

\end{enumerate}

\section{Related Work}

\paragraph{Biomedical claim verification.}
SciFact \cite{wadden-etal-2020-fact} formalized verification
with 1,409 expert-written claims annotated as
\textsc{supports}, \textsc{refutes}, or NEI. HealthVer
\cite{sarrouti-etal-2021-evidence} extends this to real-world
health queries, while PubHealth \cite{kotonya2020pubhealth}
targets public health misinformation. Most systems treat
verification as NLI problem and fine-tune encoders such as SciBERT
\cite{beltagy2019scibert}, BioBERT \cite{lee2020biobert}, or
DeBERTa \cite{he2021deberta}; MultiVerS \cite{wadden2022multivers}
advanced SciFact state of the art via full-document modeling.
\citet{kosprdic2024scientific} showed DeBERTa achieves
88\% F1 on SciFact, outperforming GPT-4 zero-shot, but
only 48\% on HealthVer OOD. We revisit this benchmark
with instruction-tuned decoder LLMs.

\paragraph{LLMs for fact verification.}
GPT-4 has been evaluated on clinical QA \cite{singhal2023medpalm, nori2023medprompt} and claim verification \cite{zheng2024judging}, achieving strong but costly zero-shot performance. Prior work on NLI artifacts \cite{gururangan-etal-2018-annotation} shows models readily exploit spurious dataset correlations.

\paragraph{Parameter-efficient fine-tuning.}
LoRA \cite{hu2022lora} injects trainable rank-decomposition matrices $\Delta W = BA$ into frozen transformer weights. QLoRA \cite{dettmers2023qlora} adds 4-bit NF4 quantization and paged optimizers, enabling single-GPU fine-tuning. QLoRA remains underexplored for biomedical NLI despite its practical fit for constrained label schemas and small datasets.

\section{Method}

\subsection{Datasets}

\textbf{SciFact} \cite{wadden-etal-2020-fact} contains 1,409 expert-written claims paired with PubMed evidence, annotated as \textsc{supports}, \textsc{refutes}, or NEI. We use an 80/20 stratified train/val split with the official dev set as test (450 examples). Label distribution: 48.8\% \textsc{supports}, 27.1\% \textsc{refutes}, 24.1\% NEI. Critically, all NEI examples have empty evidence
fields by construction. Following standard practice
for the claim verification task \cite{wadden-etal-2020-fact},
we use the annotated evidence sentences as model
input rather than full abstracts; NEI claims have
no annotated evidence by definition, resulting in
empty evidence fields.

\textbf{HealthVer} \cite{sarrouti-etal-2021-evidence} provides 14,330 evidence--claim pairs from real-world health queries verified against PubMed, with official splits (10,590/1,917/1,823). Unlike SciFact, NEI in HealthVer requires reasoning over present-but-inconclusive evidence, eliminating the absence shortcut. Label distribution: 35.7\% \textsc{supports}, 22.8\% \textsc{refutes}, 41.5\% NEI.

Both datasets use unified (claim, evidence, label) triples with deterministic label normalization (Appendix~\ref{app:normalization}). For controlled bidirectional experiments, we sample 1,008 HealthVer training examples to match SciFact training size exactly, isolating dataset structure effects from data quantity.

\subsection{Models}

\paragraph{Zero-shot baselines.} We run GPT-4o and GPT-5 at temperature 0, minimal reasoning with no chain-of-thought prompting (Appendix~\ref{app:prompts}).

\paragraph{QLoRA fine-tuned models.} We fine-tune Phi-3-mini-4k-instruct (3.8B) \cite{abdin2024phi3}, Qwen2.5-3B-Instruct \cite{qwen2025}, and Mistral-7B-Instruct-v0.3 \cite{jiang2023mistral} models. Each model is loaded in 4-bit NF4 quantization with LoRA adapters ($r{=}16$, $\alpha{=}32$) on all attention and feed-forward layers. We perform supervised fine-tuning with lr $2{\times}10^{-4}$, cosine schedule, 3 epochs and AdamW 8-bit. We select hyperparameters through grid search (Appendix~\ref{app:sweep}). We train multiple models across SciFact (1,008) and HealthVer subset (1,008) separately.
Full config is added in Appendix~\ref{app:hyperparams}.
\paragraph{Encoder baseline.} We fine-tune BioLinkBERT-base \cite{yasunaga2022linkbert} with a three-class classification head on SciFact, HealthVer subset (1,008), and HealthVer full (10,590) to provide matched and ceiling comparisons.

\subsection{Evaluation}
We report macro-averaged F1, accuracy, and per-class F1 for all three labels. Every
model trained on SciFact is evaluated on both SciFact
(in-domain) and HealthVer (out-of-domain) test sets,
and vice versa, enabling direct bidirectional comparison
under controlled conditions.

\section{Experiments and Results}

\subsection{In-Domain Performance}
\label{sec:indomain}

Table~\ref{tab:indomain} shows in-domain results for all models.

\begin{table}[t]
\centering
\small
\setlength{\tabcolsep}{3pt}
\begin{tabular}{@{}llccccc@{}}
\toprule
\textbf{Model} & \textbf{Train} & \textbf{Acc} & \textbf{F1} & \textbf{SUP} & \textbf{REF} & \textbf{NEI} \\
\midrule
\multicolumn{7}{@{}l}{\textit{SciFact test set}} \\
GPT-4o      & ---        & 85.8 & 85.6 & 86.8 & 87.7 & 82.3 \\
GPT-5       & ---        & 76.9 & 77.9 & 75.1 & \textbf{89.1} & 69.5 \\
BioLinkBERT & SF         & 87.1 & 87.5 & \textbf{86.8} & 75.6 & 100.0 \\
Phi-3-mini  & SF         & 86.2 & 86.4 & 86.0 & 73.3 & 100.0 \\
Qwen2.5-3B  & SF         & 85.3 & 86.8 & 82.4 & 78.0 & 100.0 \\
\textbf{Mistral-7B}  & SF         & \textbf{87.6} & \textbf{88.4} & 86.5 & 78.6 & 100.0 \\
\midrule
\multicolumn{7}{@{}l}{\textit{HealthVer test set (1,008-example training subset)}} \\
GPT-4o      & ---        & 56.0 & 53.2 & 39.7 & 56.8 & 63.1 \\
GPT-5       & ---        & 50.4 & 42.4 & 15.9 & 49.4 & 61.9 \\
BioLinkBERT & HV$_{sub}$ & 65.5 & 62.4 & 62.7 & 47.7 & 76.6 \\
Phi-3-mini  & HV$_{sub}$ & \textbf{66.1} & 65.1 & 64.1 & 59.8 & 71.3 \\
Qwen2.5-3B  & HV$_{sub}$ & 57.0 & 53.6 & 46.9 & 49.6 & 64.5 \\
\textbf{Mistral-7B}  & HV$_{sub}$ & 66.0 & \textbf{65.2} & \textbf{71.1} & \textbf{61.7} & 62.7 \\
\midrule
\multicolumn{7}{@{}l}{\textit{HealthVer test set (full training set, reference ceiling)}} \\
BioLinkBERT & HV$_{full}$& 82.7 & 81.9 & 81.8 & 77.4 & 86.3 \\
\bottomrule
\end{tabular}
\caption{In-domain results. SF = SciFact (1,008). HV$_{sub}$ = HealthVer 1,008-sample subset. HV$_{full}$ = full HealthVer training set (10,590), shown as reference ceiling. Metrics are percentages.}
\label{tab:indomain}
\end{table}

\paragraph{SciFact.} All three QLoRA models surpass GPT-4o macro-F1 on only 1,008 training examples. Mistral-7B achieves 88.4\%, beating GPT-4o (85.6\%) by 2.8 points and BioLinkBERT (87.5\%) by 0.9 points. McNemar's test confirms statistical indistinguishability
between Mistral and GPT-4o ($p{=}0.46$) and across
QLoRA models ($p{=}0.54$). GPT-4o holds an edge on \textsc{refutes}, reflecting broad pre-training for detecting subtle directional contradictions. GPT-5 zero-shot achieves only 77.9\% macro-F1 on SciFact, below GPT-4o and all fine-tuned
models, suggesting newer proprietary models do not automatically improve on structured verification tasks. Moreover, we discuss the mechanism behind perfect NEI score in Section~\ref{sec:shortcut}.

\paragraph{HealthVer.} Fine-tuning on just 1,008
HealthVer examples, Mistral-7B QLoRA achieves 65.2\% macro-F1, surpassing
GPT-4o, GPT-5, and BioLinkBERT (53.2\%, 42.4\%, 62.4\%). It
demonstrates that QLoRA-adapted decoders beat both the
proprietary models and the encoder approach on
real-world health queries with minimal data. The
full-training BioLinkBERT ceiling (81.9\%) shows that
adding 9,582 examples gains 19.5 macro-F1 points,
confirming data quantity effects are real but do not
close the architectural gap at matched scale. Beyond
performance, decoder models allow free-form explanation, zero-shot prompting, and do not require task-specific classification heads.

\subsection{The SciFact NEI Structural Shortcut}
\label{sec:shortcut}

High NEI F1 achieved by fine-tuned models on SciFact is not genuine epistemic reasoning. We find that all NEI examples have empty evidence fields, while every \textsc{supports} and \textsc{refutes} example contains evidence. The label is perfectly separable from evidence length alone, without reading the claim content. Full stats in Appendix~\ref{app:data_stats}, Table~\ref{tab:shortcut_stats}.

SciFact assigns NEI when no cited evidence exists,
inadvertently creating a structural signal any
expressive model will learn. Zero-shot GPT-4o attempts genuine reasoning on NEI instances. This shortcut inflates macro-F1 for all
fine-tuned models and has gone unreported in prior work.

\subsection{Bidirectional Out-of-Domain Generalization}
\label{sec:ood}

Table~\ref{tab:ood} presents the full cross-dataset evaluation.

\begin{table}[t]
\centering
\small
\setlength{\tabcolsep}{3pt}
\begin{tabular}{@{}llccccc@{}}
\toprule
\textbf{Model} & \textbf{Train$\to$Test} & \textbf{Acc} & \textbf{F1} & \textbf{SUP} & \textbf{REF} & \textbf{NEI} \\
\midrule
\multicolumn{7}{@{}l}{\textit{SciFact-trained $\to$ HealthVer}} \\
GPT-4o      & ---$\to$HV       & 56.0 & 53.2 & 39.7 & 56.8 & 63.1 \\
GPT-5       & ---$\to$HV       & 50.4 & 42.4 & 15.9 & 49.4 & 61.9 \\
BioLinkBERT & SF$\to$HV        & 40.3 & 31.5 & 53.5 & 40.9 &  0.3 \\
Phi-3-mini  & SF$\to$HV        & 42.8 & 36.0 & 55.1 & 44.9 &  8.0 \\
Qwen2.5-3B  & SF$\to$HV        & 45.4 & 43.2 & 54.7 & 47.3 & 27.5 \\
\textbf{Mistral-7B}  & SF$\to$HV        & \textbf{48.9} & \textbf{44.4} & \textbf{60.1} & \textbf{52.9} & \textbf{20.1} \\
\midrule
\multicolumn{7}{@{}l}{\textit{HealthVer-trained $\to$ SciFact (matched 1,008 examples)}} \\
GPT-4o      & ---$\to$SF       & 85.8 & 85.6 & 86.8 & 87.7 & 82.3 \\
GPT-5       & ---$\to$SF       & 76.9 & 77.9 & 75.1 & 89.1 & 69.5 \\
BioLinkBERT & HV$_{sub}$$\to$SF& 61.1 & 53.3 & 70.8 & 21.9 & 67.1 \\
Phi-3-mini  & HV$_{sub}$$\to$SF& 63.3 & 59.2 & 71.3 & 43.3 & 63.1 \\
Qwen2.5-3B  & HV$_{sub}$$\to$SF& 41.3 & 37.9 & 39.1 & 27.8 & 46.7 \\
\textbf{Mistral-7B}  & HV$_{sub}$$\to$SF& \textbf{72.4} & \textbf{69.3} & \textbf{77.8} & \textbf{55.7} & \textbf{74.3} \\
\midrule
\multicolumn{7}{@{}l}{\textit{HealthVer full $\to$ SciFact (reference ceiling, 10,590 examples)}} \\
BioLinkBERT & HV$_{full}$$\to$SF& 68.9 & 60.8 & 77.9 & 29.1 & 75.3 \\
\bottomrule
\end{tabular}
\caption{Bidirectional OOD results. SF = SciFact; HV = HealthVer. Top: SciFact-trained models on HealthVer test. Middle: HealthVer subset (1,008)-trained models on SciFact test. Bottom: BioLinkBERT trained on full HealthVer (reference). GPT-4o shown for reference}
\label{tab:ood}
\end{table}

\paragraph{SciFact $\to$ HealthVer.} All SciFact-trained models suffer catastrophic degradation as macro-F1 drops from 86-88\% in-domain to 36-44\% OOD. NEI collapse is most severe as HealthVer NEI has non-empty, topically relevant evidence and the absence shortcut does not fire. Table~\ref{tab:ood} shows the results. Qwen2.5-3B is the most NEI-resilient SciFact-trained model, suggesting broader pre-training provides some generalization advantage.

\paragraph{HealthVer $\to$ SciFact.} Reverse
direction tells a strikingly different story. Mistral-7B
trained on just 1,008 HealthVer examples achieves 69.3\%
macro-F1 OOD on SciFact with 74.3\% NEI F1, far above
any SciFact-trained model on HealthVer. Remarkably, this
outperforms BioLinkBERT trained on the full HealthVer
set (60.8\%), despite using 10$\times$ less training
data. BioLinkBERT at matched size shows meaningful
transfer and achieves 53.3\% macro-F1 and 67.1\% NEI F1. Together, these results demonstrate that
training on structurally sound data enables robust cross-domain transfer and model architecture
matters more than data quantity.

\paragraph{Asymmetric OOD generalization.} Symmetric
degradation would indicate distributional shift; instead,
collapse is strictly one-directional. The asymmetry
holds at matched training sizes (1,008 examples), ruling
out data quantity. Models trained on genuine epistemic
reasoning (HealthVer) transfer; models trained on a
structural proxy (SciFact) do not.

\paragraph{Per-class analysis.} \textsc{Refutes} F1
degrades most severely OOD in both directions:
BioLinkBERT (HealthVer$_{full}$$\to$SciFact) achieves
only 29.1\% despite strong \textsc{supports} (77.9\%)
and NEI (75.3\%) transfer. Directional contradiction
detection requires domain-specific reasoning that
generalizes poorly across biomedical claim types. This is a
critical gap given that refutation detection is
essential for any clinical task.

\paragraph{GPT-4o and GPT-5: paradoxical OOD behavior.}
GPT-4o achieves the highest HealthVer NEI F1 (63.1\%)
among zero-shot models despite the lowest SciFact NEI
F1 (82.3\%). Its in-domain weakness reflects genuine
uncertainty reasoning that transfers. GPT-5 shows more
pronounced NEI over-prediction: 95.5\% NEI recall on
SciFact but only 8.8\% SUP recall on HealthVer,
suggesting over-cautious RLHF training that predicts
NEI regardless of evidence. In-domain scores are poor
proxies.

\subsection{Cost Analysis}
\label{sec:cost}

\begin{table}[t]
\centering
\small
\begin{tabular}{@{}lrrr@{}}
\toprule
\textbf{Model} & \textbf{Cost/1K} & \textbf{Fine-tune} & \textbf{vs.\ GPT-4o} \\
\midrule
GPT-4o (API)    & \$1.3000 & ---    & 1.0$\times$ \\
GPT-5 (API)     & \$0.5750 & ---    & 2.4$\times$ cheaper \\
\midrule
Phi-3 (QLoRA)   & \$0.0292 & \$0.35 & 44.5$\times$ cheaper \\
Qwen2.5 (QLoRA) & \$0.0292 & \$0.35 & 44.5$\times$ cheaper \\
Mistral (QLoRA) & \$0.0292 & \$0.35 & 44.5$\times$ cheaper \\
BioLinkBERT     & \$0.0292 & \$0.17 & 44.5$\times$ cheaper \\
\bottomrule
\end{tabular}
\caption{Inference cost per 1,000 preds and one-time fine-tuning cost (T4 GPU). API pricing as of early 2026.}
\label{tab:cost}
\end{table}

GPT-4o costs \$1.30 per 1,000 predictions versus \$0.03 for local open models (44.5$\times$ drop). The \$0.35 one-time fine-tuning cost amortizes to negligible overhead, making fine-tuned small models strictly preferable on both performance and cost grounds. For OOD
tasks, training data choice is critical.
SciFact-trained models collapse OOD while
HealthVer-trained models transfer robustly, achieving
strong cross-domain performance.

\subsection{Qualitative Error Analysis}
\label{sec:error}

We examined 48 cases where Mistral-7B QLoRA is correct
and GPT-4o is wrong, and 40 reverse cases, on the
SciFact test set.

\textbf{Fine-tuning wins.} Mistral-7B outperforms
GPT-4o on NEI: for ``Statins increase blood
cholesterol,'' GPT-4o predicts \textsc{refutes} from
pharmacological knowledge while Mistral correctly
withholds judgment given no retrieved evidence.
Fine-tuning teaches models to attend to evidence
rather than prior knowledge.

\textbf{GPT-4o wins.} GPT-4o retains an edge on
directional \textsc{refutes}: for ``The risk of male
prisoners harming themselves is ten times that of
female prisoners,'' the evidence states the opposite
direction. GPT-4o correctly predicts \textsc{refutes};
Mistral latches onto ``ten times'' without detecting
the subject inversion. Mistral nonetheless closes this
gap vs.\ Phi-3-mini (11 GPT-4o wins vs.\ 29),
suggesting larger capacity helps with contradiction detection.

\textbf{Shared failure modes.} Both models err on
claims where evidence describes a related but
experimentally distinct condition, and on logical
negations introduced by experimental targeting. The
annotation ambiguity is documented in SciFact
inter-annotator studies \cite{wadden-etal-2020-fact}.

\section{Conclusion}

Fine-tuning small open-weight LLMs via QLoRA on just
1,008 examples surpasses both GPT-4o and GPT-5 on
SciFact and HealthVer at a fractional cost, establishing
parameter-efficient fine-tuning as a practical
alternative to APIs. Through extensive OOD evaluation, we show both the promise and limits of this approach: HealthVer-trained
models transfer robustly to SciFact, outperforming
BioLinkBERT trained on 10$\times$ more data, while
SciFact-trained models collapse cross-domain due to a
structural artifact in SciFact. \textsc{Refutes} remains the hardest class to transfer in both directions, pointing to cross-domain
contradiction detection as an open challenge. Our
findings motivate structural auditing of biomedical NLI
benchmarks and bidirectional evaluation as standard
practice.

\section*{Limitations}
Our bidirectional experiments use matched 1,008-example
training sets, but scaling to full HealthVer data would
further improve in-domain performance, as evidenced by
BioLinkBERT's 81.9\% ceiling vs.\ 62.4\% at matched
size. We evaluated GPT-5 as an additional zero-shot
baseline; its lower performance (77.9\% SciFact,
42.4\% HealthVer) confirms that QLoRA fine-tuning
advantages hold against the latest proprietary systems.
Cost estimates reflect T4/A100 and API pricing as of
early 2026. We evaluate English claims only; whether
analogous structural shortcuts exist in PubHealth,
MedNLI, or other benchmarks remains an open question.
Data augmentation and debiasing strategies for
improving cross-domain robustness are left for future
work.

\bibliography{custom}

\clearpage
\appendix

\section{Label Normalization}
\label{app:normalization}
Refer to Table \ref{tab:canonical}.
\begin{table}[h]
\centering
\small
\begin{tabular}{@{}ll@{}}
\toprule
\textbf{Raw output variants} & \textbf{Label} \\
\midrule
SUPPORTS, SUPPORT, SUPPORTED, & \multirow{2}{*}{\textsc{supports}} \\
\quad ENTAILMENT & \\
\midrule
REFUTES, REFUTE, REFUTED, & \multirow{2}{*}{\textsc{refutes}} \\
\quad CONTRADICT(S) & \\
\midrule
NEI, NOT\_ENOUGH\_INFO, & \multirow{2}{*}{\textsc{nei}} \\
\quad NOT ENOUGH INFO, NEUTRAL & \\
\midrule
\textit{(unrecognized)} & \textsc{nei} (fallback) \\
\bottomrule
\end{tabular}
\caption{Label normalization mapping}
\label{tab:canonical}
\end{table}

\section{Prompt Templates}
\label{app:prompts}

All instruction-tuned LLMs use the following prompt. Phi-3-mini uses ChatML format; Mistral-7B and Qwen2.5 use \texttt{[INST]/[/INST]} format.

\begin{quote}
\small
\texttt{System: You are a biomedical claim verification expert. Given a claim and an evidence passage from a scientific abstract, determine whether the evidence SUPPORTS the claim, REFUTES the claim, or whether there is NOT ENOUGH INFO. Respond with exactly one of: SUPPORTS, REFUTES, NEI.}\\[4pt]
\texttt{User: Claim: \{claim\}}\\
\texttt{Evidence: \{evidence\}}\\
\texttt{What is the verdict?}\\[4pt]
\texttt{Assistant: \{label\}}
\end{quote}

For GPT-4o, only the system and user turns are used at inference. For fine-tuning, the full exchange including the gold label is the SFT training target. For SciFact NEI examples, the evidence field is empty (\texttt{Evidence: }), the structural signal described in Section~\ref{sec:shortcut}. We run GPT-4o with temperature 0 and GPT-5 with minimal reasoning.

\section{Dataset Statistics}
\label{app:data_stats}

Tables~\ref{tab:data_stats}, \ref{tab:label_dist} present dataset split sizes and label distributions. Tables~\ref{tab:dataset_examples}, \ref{tab:shortcut_stats} provide representative input examples and evidence field
statistics respectively
\begin{table}[h]
\centering
\small
\begin{tabular}{@{}lrrrr@{}}
\toprule
\textbf{Dataset} & \textbf{Train} & \textbf{Val} & \textbf{Test} & \textbf{Total} \\
\midrule
SciFact              & 1,008  & 253   & 450   & 1,711  \\
HealthVer (subset)   & 1,008  & ---   & 1,823 & ---    \\
HealthVer (full)     & 10,590 & 1,917 & 1,823 & 14,330 \\
\bottomrule
\end{tabular}
\caption{Dataset splits. SciFact uses stratified 80/20 train/val split; official dev set as test. HealthVer subset sampled with random\_state=42 to match SciFact training size.}
\label{tab:data_stats}
\end{table}

\begin{table}[h]
\centering
\small
\begin{tabular}{@{}lrrr@{}}
\toprule
\textbf{Dataset (train)} & \textbf{SUP\%} & \textbf{REF\%} & \textbf{NEI\%} \\
\midrule
SciFact        & 48.8 & 27.1 & 24.1 \\
HealthVer      & 35.7 & 22.8 & 41.5 \\
\bottomrule
\end{tabular}
\caption{Label distributions. HealthVer's higher NEI proportion and structurally distinct NEI definition contribute to the generalization asymmetry.}
\label{tab:label_dist}
\end{table}

\begin{table}[h]
\centering
\small
\begin{tabular}{@{}p{1.1cm}p{1.8cm}p{2.9cm}p{0.6cm}@{}}
\toprule
\textbf{Dataset} & \textbf{Claim} & \textbf{Evidence} &
\textbf{Label} \\
\midrule
SciFact & Statins increase blood cholesterol. &
\textit{(empty)} & NEI \\
SciFact & Metformin reduces HbA1c in diabetic patients. &
Metformin significantly reduced HbA1c levels compared to placebo... & SUP \\
SciFact & Aspirin prevents colorectal cancer. &
Aspirin use was associated with reduced risk... & REF \\
\midrule
HealthVer & Vitamin C prevents the common cold. &
Some studies suggest marginal effects on duration but evidence remains inconclusive... & NEI \\
HealthVer & Exercise reduces depression symptoms. &
Aerobic exercise significantly reduced depressive symptoms across trials... & SUP \\
\bottomrule
\end{tabular}
\caption{Representative input examples from SciFact
and HealthVer. SciFact NEI examples have empty
evidence fields by construction; HealthVer NEI
examples contain present-but-inconclusive evidence,
eliminating the absence shortcut. Evidence text
truncated for brevity. This structural difference
explains the asymmetric OOD generalization reported
in Section~\ref{sec:ood}.}
\label{tab:dataset_examples}
\end{table}

\FloatBarrier
\begin{table}[h]
\centering
\small
\begin{tabular}{@{}lrrr@{}}
\toprule
\textbf{Label} & \textbf{Count} & \textbf{Mean chars} & \textbf{Zero-length} \\
\midrule
NEI               & 243 & 0.0   & 243/243 (100\%) \\
\textsc{supports} & 492 & 214.5 & 0/492 (0\%)     \\
\textsc{refutes}  & 273 & 227.4 & 0/273 (0\%)     \\
\bottomrule
\end{tabular}
\caption{Evidence field length by label in SciFact training split. All NEI examples have empty evidence; no SUP/REF example does. This holds in val and test splits as well.}
\label{tab:shortcut_stats}
\end{table}

\section{Hyperparameter Grid Search}
\label{app:sweep}

We perform multi configuration grid search over $r \in \{8, 16, 32\}$ and lr $\in \{10^{-4}, 2{\times}10^{-4}, 5{\times}10^{-4}\}$ using Mistral-7B on SciFact validation set. Selected config ($r{=}16$, lr$=2{\times}10^{-4}$) in bold.

\begin{table}[h]
\centering
\small
\begin{tabular}{@{}ccr@{}}
\toprule
$r$ & \textbf{LR} & \textbf{Val Macro-F1} \\
\midrule
\textbf{16} & $\mathbf{2{\times}10^{-4}}$ & \textbf{88.4} \\
8  & $5\times10^{-4}$ & 87.4 \\
16 & $5\times10^{-4}$ & 85.7 \\
32 & $5\times10^{-4}$ & 84.6 \\
8  & $2\times10^{-4}$ & 81.3 \\
32 & $2\times10^{-4}$ & 79.0 \\
16 & $1\times10^{-4}$ & 77.8 \\
32 & $1\times10^{-4}$ & 75.7 \\
8  & $1\times10^{-4}$ & 74.5 \\
\bottomrule
\end{tabular}
\caption{Hyperparameter grid search results on SciFact
validation set (Mistral-7B). Selected configuration in bold. Higher learning rates generally outperform lower ones; $r{=}16$ at lr$=2{\times}10^{-4}$ achieves the best validation macro-F1 of 88.4\%.}
\label{tab:sweep}
\end{table}

\section{Hyperparameter Configuration}
\label{app:hyperparams}
We list down the hyperparameters used for fine-tuning the LLMs in Table \ref{tab:hyperparams}.

\begin{table}[h]
\centering
\small
\begin{tabular}{@{}ll@{}}
\toprule
\textbf{Hyperparameter} & \textbf{Value} \\
\midrule
LoRA rank ($r$)           & 16 \\
LoRA alpha ($\alpha$)     & 32 \\
LoRA dropout              & 0.05 \\
Quantization              & NF4, double quantization \\
Learning rate             & $2 \times 10^{-4}$ \\
LR schedule               & Cosine, warmup ratio 0.05 \\
Effective batch size      & 16 (batch 4, grad.\ accum.\ 4) \\
Max sequence length       & 1,024 tokens \\
Training epochs           & 3 \\
Precision                 & bf16 (fp16 fallback on T4) \\
Optimizer                 & Paged AdamW 8-bit \\
Checkpoint selection      & Best validation macro-F1 \\
Trainable params (Phi-3)        & 8.9M / 3.83B (0.23\%) \\
Trainable params (Mistral/Qwen) & 41.9M / 7.29B (0.58\%) \\
\midrule
\multicolumn{2}{@{}l}{\textit{LoRA target modules:}} \\
\multicolumn{2}{@{}l}{\texttt{q\_proj, k\_proj, v\_proj, o\_proj,}} \\
\multicolumn{2}{@{}l}{\texttt{gate\_proj, up\_proj, down\_proj}} \\
\bottomrule
\end{tabular}
\caption{QLoRA configuration, identical across all architectures and datasets.}
\label{tab:hyperparams}
\end{table}

\section{Training Efficiency}
\label{app:hardware}
We measure training efficiency as a function of
training time and cost. All models were trained on
Google Colab Pro. Phi-3-mini, Qwen2.5-3B, and
BioLinkBERT were trained on a T4 GPU (16 GB);
Mistral-7B required an A100 (40 GB) due to memory
requirements at 7B scale. Times are reported for
SciFact (1,008 examples, 3 epochs); HealthVer
subset training is approximately equivalent.
The detailed breakdown is in Table~\ref{tab:hardware}.

\begin{table}[h]
\centering
\small
\begin{tabular}{@{}llrr@{}}
\toprule
\textbf{Model} & \textbf{GPU} & \textbf{Time} & \textbf{Cost} \\
\midrule
Phi-3-mini  & T4 (16 GB)   & $\sim$50 min & \$0.15 \\
Qwen2.5-3B  & T4 (16 GB)   & $\sim$70 min & \$0.21 \\
Mistral-7B  & A100 (40 GB) & $\sim$35 min & \$0.35 \\
BioLinkBERT & T4 (16 GB)   & $\sim$15 min & \$0.04 \\
\bottomrule
\end{tabular}
\caption{Training times and costs per model (SciFact, 1,008 examples, 3 epochs). T4: $\approx$\$1.76 CU/hr; A100: $\approx$\$10--15 CU/hr.}
\label{tab:hardware}
\end{table}

\section{McNemar's Test Details}
\label{app:mcnemar}

McNemar's test statistic: $\chi^2 = (b_{01} - b_{10})^2 / (b_{01} + b_{10})$, where $b_{01}$ = model A correct, B wrong; $b_{10}$ = model B correct, A wrong. Evaluated on SciFact test set (450 examples).

\begin{table}[h]
\centering
\small
\begin{tabular}{@{}llrrrr@{}}
\toprule
\textbf{Model A} & \textbf{Model B} & $b_{01}$ & $b_{10}$ & $\chi^2$ & $p$ \\
\midrule
Mistral QLoRA & GPT-4o        & 48 & 40 & 0.56 & 0.46 \\
Mistral QLoRA & Phi-3 QLoRA   & 36 & 30 & 0.38 & 0.54 \\
\bottomrule
\end{tabular}
\caption{McNemar's test on SciFact test set. Neither
comparison is significant at $p < 0.05$.}
\label{tab:mcnemar}
\end{table}

\section{Qualitative Error Analysis: Extended Examples}
\label{app:error}

\begin{center}
\small
\setlength{\tabcolsep}{3pt}
\begin{tabular}{@{}p{4.5cm}p{0.6cm}p{1.2cm}p{1.2cm}@{}}
\toprule
\textbf{Claim} & \textbf{Gold} & \textbf{Mistral} &
\textbf{GPT-4o} \\
\midrule
\multicolumn{4}{@{}l}{\textit{Pattern A: Fine-tuned wins on NEI}} \\
Statins increase blood cholesterol. & NEI &
\checkmark~NEI & $\times$~REF \\
Venules have larger lumen than arterioles. & NEI &
\checkmark~NEI & $\times$~SUP \\
\midrule
\multicolumn{4}{@{}l}{\textit{Pattern B: GPT-4o wins on REFUTES}} \\
Male prisoner self-harm risk is 10$\times$ female. &
REF & $\times$~SUP & \checkmark~REF \\
Nicotine combo therapy yields significantly higher
abstinence. & REF & $\times$~SUP & \checkmark~REF \\
\midrule
\multicolumn{4}{@{}l}{\textit{Pattern C: Mistral wins on REFUTES (new vs.\ Phi-3)}} \\
NIV use should be decreased for poor responders. &
REF & \checkmark~REF & $\times$~NEI \\
Tirasemtiv has no effect on fast-twitch muscle. &
REF & \checkmark~REF & $\times$~NEI \\
\midrule
\multicolumn{4}{@{}l}{\textit{Pattern D: Both wrong}} \\
TNFAIP3 is a tumor suppressor in glioblastoma. &
REF & $\times$~SUP & $\times$~SUP \\
Low nucleosome occupancy correlates with low
methylation. & REF & $\times$~SUP & $\times$~SUP \\
\bottomrule
\end{tabular}
\captionof{table}{Representative error examples.
REF = \textsc{refutes}, SUP = \textsc{supports}.}
\label{tab:error_examples}
\end{center}

\textbf{A.} Fine-tuned models withhold judgment
without evidence; GPT-4o applies world knowledge.

\textbf{B.} GPT-4o detects subject inversions and
non-significant p-values; Mistral closes this gap
vs.\ Phi-3-mini (11 wins vs.\ 29).

\textbf{C.} Mistral infers directional implications
from indirect evidence where GPT-4o hedges to NEI.

\textbf{D.} Both models fail on logical negations and
directional correlation signs \cite{wadden-etal-2020-fact}.

\section{Confusion Matrices}
\label{app:confusion}

\begin{figure}[h]
\centering
\includegraphics[width=0.98\columnwidth]{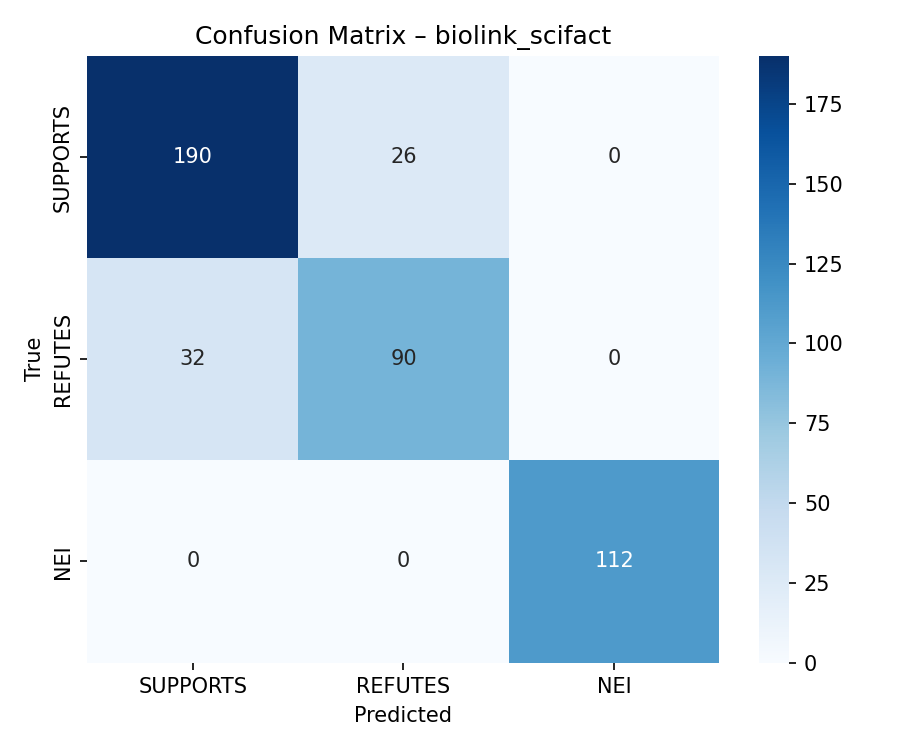}\\[6pt]
\includegraphics[width=0.98\columnwidth]{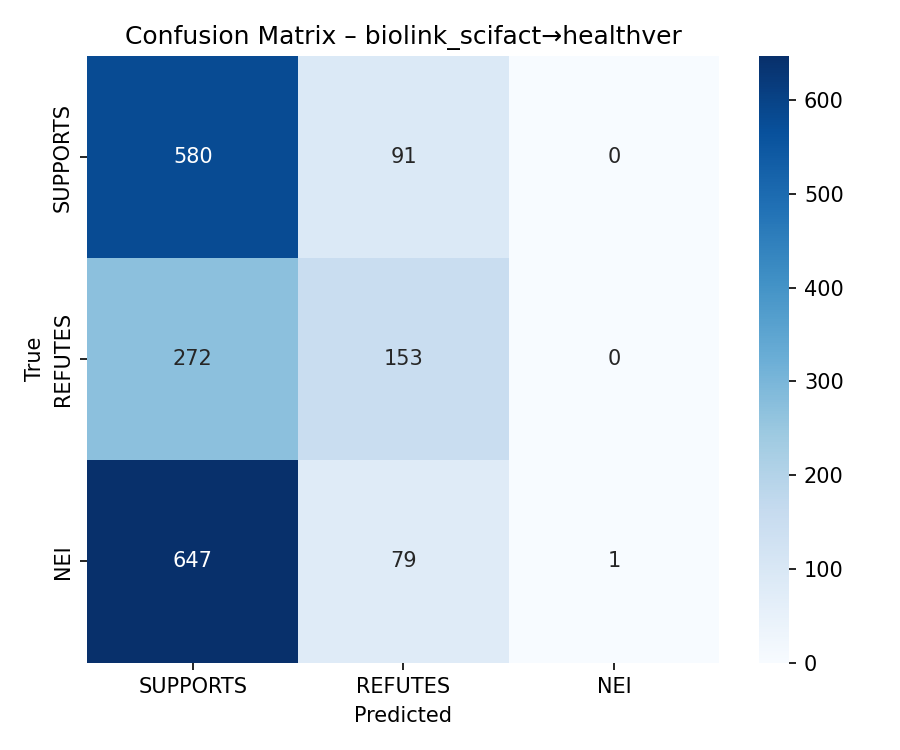}
\caption{Confusion matrices for BioLinkBERT trained on
SciFact, evaluated in-domain (top) and OOD on HealthVer
(bottom). In-domain NEI is perfectly classified (112/112);
OOD NEI is nearly never predicted (1/727), with true NEI
examples overwhelmingly misclassified as \textsc{supports}.
This is the behavioral signature of structural shortcut
learning.}
\label{fig:cm_biolink}
\end{figure}

Figure~\ref{fig:cm_biolink} visualizes the NEI shortcut
directly. BioLinkBERT trained on SciFact achieves perfect
NEI separation in-domain (left). The NEI column is
completely clean with zero misclassifications. When
evaluated on HealthVer OOD (right), the model predicts
NEI exactly once across 727 true NEI examples, instead
routing all NEI instances to \textsc{supports} (647) or
\textsc{refutes} (79). This collapse is the behavioral
signature of shortcut learning: the absence signal that
perfectly predicted NEI in SciFact is absent in HealthVer,
and the model has learned nothing else.

\end{document}